# Make it more specific: A novel uncertainty based airway segmentation application on 3D U-Net and its variants

Shiyi Wang
National Heart and Lung Institute,
Imperial College London
London, United Kingdom
s.wang22@imperial.ac.uk

Yang Nan
National Heart and Lung Institute,
Imperial College London
London, United Kingdom
y.nan20@imperial.ac.uk

Felder Federico N
National Heart and Lung Institute,
Imperial College London
London, United Kingdom
f.felder@imperial.ac.uk

Sheng Zhang
National Heart and Lung Institute,
Imperial College London
London, United Kingdom
sheng.zhang@imperial.ac.uk

Walsh Simon L F
Royal Brompton Hospital
National Heart and Lung
Institute, Imperial College
London
London, United Kingdom
s.walsh@imperial.ac.uk

Guang Yang
Cardiovascular Research Centre,
Royal Brompton Hospita
National Heart and Lung
Institute, Imperial College
London
London, United Kingdom
g.yang@imperial.ac.uk

*Abstract*—Each medical segmentation task should be considered with a specific AI algorithm based on its scenario so that the most accurate prediction model can be obtained. The most popular algorithms in medical segmentation, 3D U-Net and its variants, can directly implement the task of lung trachea segmentation, but its failure to consider the special tree-like structure of the trachea suggests that there is much room for improvement in its segmentation accuracy. Therefore, a research gap exists because a great amount of state-of-the-art DL algorithms are vanilla 3D U-Net structures, which do not introduce the various performance-enhancing modules that come with special natural image modality in lung airway segmentation. In this paper, we proposed two different network structures Branch-Level U-Net (B-UNet) and Branch-Level CE-UNet (B-CE-UNet) which are based on U-Net structure and compared the prediction results with the same dataset. Specially, both of the two networks add branch loss and central line loss to learn the feature of fine branch endings of the airways. Uncertainty estimation algorithms are also included to attain confident predictions and thereby, increase the overall trustworthiness of our whole model. In addition, predictions of the lung trachea based on the maximum connectivity rate were calculated and extracted during post-processing for segmentation refinement and pruning.

## I. Introduction

The pulmonary airways are the body's only means of gas exchange with the outside world, so the anatomical information obtained from airway segmentation can be used to diagnose respiratory diseases. Lung related diseases such as chronic obstructive pulmonary disease (COPD) [1], specially, idiopathic pulmonary fibrosis (IPF) is the archetypal, most common, and lethal progressive fibrotic lung disease with a median survival of 3 years, and 5-year survival of 25%; worse than most cancers. For instance, a key challenge facing clinicians managing patients with IPF is that is currently not possible to reliably predict which patients will progress and which will remain stable. This makes management difficult, expensive and risky for patients as they may end up having a biopsy which has a 2% 30 day in hospital mortality. High-resolution computed tomography (HRCT) is the cornerstone imaging technique for diagnosis and prognostication in patients with suspected lung disease and is routinely performed in all patients with suspected lung disease. HRCT is a non-invasive investigation allowing visual assessment of disease morphology and extent. In several studies, COPD is characterized by hyperplasia and thickening of the airway walls, which can as a biomarker of mortality prognosis [2]–[4]. Similarly, severity of traction bronchiectasis, which represents abnormal dilatation of the airways caused by the surrounding fibrotic lung tissue, has been reported as a strong predictor of mortality in several fibrotic lung disease subsets including idiopathic fibrotic lung disease, connective tissue disease (CTD) related fibrotic lung disease and chronic fibrotic hypersensitivity pneumonitis [5]. In addition to these, the airway tree subdivides the lung tissue and tasks such as lobes and segments also need to be dependent on the outcome of the segmentation of the trachea [6]. An automated lung airway segmentation model can be used to assist physicians in the diagnosis of IPF and patient prognosis in the clinical field for the above lung diseases associated with abnormal lung airway morphology.

Deep learning (DL) has been successfully applied to airway segmentation on HRCT [7]. Separating abnormal airways, especially the tiny lung tracheal ends, from the surrounding reticular/foveal structures (the hallmark pattern of fibrosis on HRCT) is particularly difficult due to the confounding effects of co-existing fibrosis. Because 2D HRCT slices may lose the significant 3D context of the whole original 3D HRCT images, the resulting predictions of 3D models should have a higher performance than 2D airway segmentation, theoretically. A 3D UNet baseline was proposed to develop different advanced DL models specific to the lung airway segmentation task. For instance, our 3D Branch-Level CE-UNet (B-CE-UNet) added artous convolution blocks, branch loss function, central line loss function and Bayesian uncertainty estimation in the proposed 3D UNet baseline for enhancing performance of lung airway segmentation. Our novel algorithms can be seamlessly integrated into existing DL models. Due to its remarkable capability to capture features in tree-like pipeline structures and intricate pipeline textures, it can be effectively applied in tasks such as vessel and airway segmentation. By leveraging these unique strengths, our algorithm enhances the accuracy of the original predictions, yielding more precise outcomes.

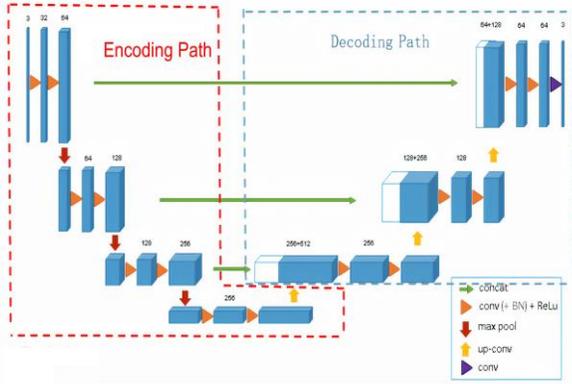

Fig. 1 The 3D U-Net structure

## II. METHODS

### A. Dataset and pre-processing

A total of 140 cases from a mix of EXACT09 dataset [8] and LIDC-IDRI [9]dataset of 3D HRCT data were used for training, validation and testing. 72 cases were used for training and 18 cases were used for validation (with training rate 0.8/0.2 of dataset), 50 cases for testing. Each case contains a pair of the original images and ground truth (GT). It is worth noting that any data used for testing or inference purposes has never been utilized during the training and validation processes. We aim to maintain the integrity and objectivity of the model evaluation.

Due to the varying volumes of 3D high-resolution computed tomography (HRCT) images, their three-dimensional arrays can be represented by dimensions as [512, 512, xxx]. However, this size proves to be too large for the input layer of a 3D network, demanding excessive GPU resources for processing. As a result, we adopted adjust window method to divide the complete HRCT images into the same 3D patches of size [128, 96, 144]. Furthermore, we employed random flip, random rotate, scale transformation and z-score normalization operations on the data to augment its diversity, enhance the model's generalization capability, and facilitate the learning of object features in different orientations.

Identify applicable funding agency here. If none, delete this text box.

### B. Bsaeline, our Branch-Level U-Net and our Branch-Level CE-UNet

**3D U-Net** [10] learns from sparse annotations and provides a dense 3D segmentation mask corresponding to 3D HRCT images. Due to the limited amount of medical data containing expert annotations and the difficulty of obtaining them, 3D U-Net can be trained with very little data to obtain excellent generalisation results. In contrast to the 2D U-Net network structure, there is an encoding path and a decoding path, each with 4 resolution levels. Each layer of the encoding path contains two $3\times3\times3$ convolutions, each followed by a ReLU layer, followed by a $2\times2\times2$ maximum pooling layer with a step size of 2 in each direction. In the decoding path, each layer contains a $2\times2\times2$ deconvolution layer with a step size of 2, followed by two $3\times3\times3$ convolutions and an Instance normalization, then followed by a RuLU layer. A final layer of $1\times1\times1$ convolution at the end of the encoder provides the original high-resolution features, which reduces the number of channels (i.e. the number of categories of labels) in the output.

Weighted softmax loss function in the network architecture allows the network to be trained using sparsely annotated data. Setting the weights of unlabelled pixels to zero allows the network to learn from only the labelled pixels and generalise to the entire stereo data. Softmax with weighted cross-entropy loss reduces the weight of the background to achieve a balanced effect of tubule and background voxels on the loss. The main purpose is to transform data of different magnitudes into the same magnitude, measured uniformly by the calculated z-score value, in order to ensure comparability between the data. z-score normalization allows the pixel values of two images to be mapped into a distribution of values that conform to a positive-terrestrial distribution with a mean of 0 and a standard deviation of 1 (taking values in the range [0,1]). It somewhat eliminates the effect of updating the model weight values due to overexposure, poor quality or noise for various reasons. Fig.1 shows the detailed structure of our 3D U-Net baseline model.

**Branch-Level U-Net (B-UNet)** based on U-Net but focus more on the tree-like structures in the airways of the lungs. We proposed B-UNet with the novel branch loss function, central line loss function, uncertainty estimation using both Monte Carlo (MC) method and inference dropout method. These methods will introduce in the following paragraph.

**Branch -Level CE-UNet (B-CE-UNet)** based on B-UNet. Inspired by the context extraction module of CE-Net architecture [11], we propose a novel branch-level specific lung airway segmentation model to address the issue of spatial information loss due to the consecutive pooling and convolutional operations in B-UNet. Between the encoder and decoder in B-UNet, a context extraction module consists of a dense atrous convolution (DAC) module and a residual multi-kernel pooling (RMP) module is implemented (See Fig.2).

The task of densely predicting detailed spatial information requires the maintenance of high-resolution feature maps in intermediate layers to improve segmentation accuracy without increasing the training cost. Therefore, a dense atrous convolution (DAC) block composed of atrous convolutions is employed, which captures more extensive and deeper semantic features by injecting four cascaded branches with multiscale atrous convolutions (Fig.3). Gradually increasing the number of dilated convolutions and adjusting the dilation rates expands the network's receptive field, enabling the capture of a wider range of contextual information and improving the model's understanding of global or long-range dependencies. The receptive field refers to the region size from which each neuron in a convolutional neural network receives input data. In each atrous branch, a 1*1*1 convolution is applied for RELU.

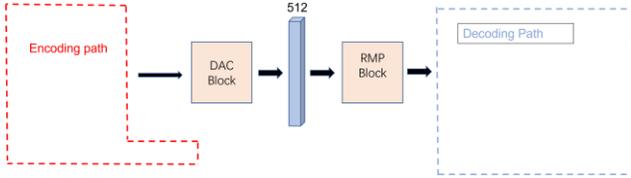

Fig. 2 The structure of proposed B-CE-UNet

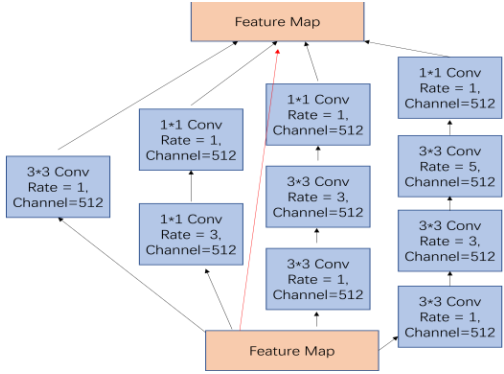

Fig. 3. DAC block. * The red arrows indicate the use of shortcut connections in DAC to mitigate the issue of gradient vanishing. A single dilated convolutional layer with a dilation rate of 1 is employed, resulting in a receptive field size of 3. Subsequently, the number of dilated convolutional layers is increased to three, with dilation rates of 1, 3, and 5, respectively, corresponding to receptive field sizes of 3, 7, and 9. Finally, the number of dilated convolutional layers is further increased to five, with dilation rates of 1, 3, 5, 7, and 9, resulting in receptive field sizes of 3, 7, 9, and 19, respectively.

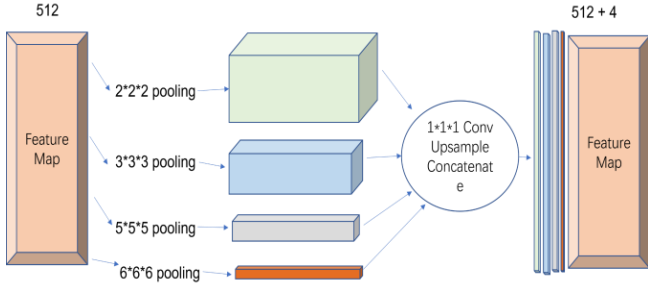

Fig. 4 The RMP block

Residual multinuclear pooling (RMP) relies on multiple effective fields of view to detect targets of different sizes. The size of the receptive field roughly determines how much contextual information is used. A general maximum pooling uses only a single pooling kernel, e.g. 2*2*2. As shown in Figure 5, the proposed RMP encodes global contextual information with four different sizes of perceptual fields: 2*2*2,3*3*3,5*5*5 and 6*6*6. The output of the four branches contains feature maps of various sizes. To reduce the dimensionality and computational cost of the weights, a 1*1*1 convolution is used after each pooling branch. The convolution layer reduces the size of the feature map to 1/N of the original size, where N denotes the number of channels in the original feature map. The low-dimensional feature map is then upsampled to obtain features of the same size as the original feature map by bilinear interpolation. Finally, the original features are combined with the upsampled featuremap.

This paper used the same and training parameters in all three models, e.g. 100 epochs, batch size = 1, learning rate of 5e-4, optimiser of Adam.

*C. Branch Loss*

Each branch can be calculated and labeled by the parent-children relationship. Due to the availability of tree parsing, it is possible to compute the branches for each segment. The intersection is obtained by multiplying the predicted branch for each segment with the corresponding ground truth (GT) branch and summing the results. The denominator is obtained by summing the branches of the ground truth for each segment. Consequently, the branch loss can be expressed as follows:

$$L_{branch\_loss} = 1 - \frac{\sum Pred_{branch\_label} * GT_{branch_{label}} + smooth}{\sum GT_{branch\_label} + smooth} \quad (1)$$

Branch_label represents the index of each branch.

*D. Central Line Loss*

The calculation method for central line loss is similar to branch loss, with the additional step of extracting skeletons from both the predicted branch and the ground truth. This is achieved by employing image erosion to extract the complete 3D airway's central line, where the airway's internal diameter is defined as a single-pixel size. Consequently, the central line loss can be expressed as follows:

$$E_{pred} = f(x) = erosion(Pred_{branch_{label}})$$

$$E_{GT} = f(x) = erosion(GT_{branch\_label})$$

$$L_{central\_line\_loss} = 1 - \frac{\sum E_{pred} * E_{GT} + smooth}{\sum E_{GT} + smooth} \quad (2)$$

Therefore, by incorporating the Dice Loss (dice similarity coefficient) and BCE Loss (binary cross entropy), the overall Loss function can be obtained:

$$L_{total} = L_{Dice} * w_1 + L_{BCE} * w_2 + L_{Branch} * w_3 + L_{Central_{Line}} * w_4 \quad (3)$$

In the context mentioned, w1, w2, w3, and w4 represent the weights multiplied with different loss functions. These weights can be tuned according to individual requirements and preferences, e.g. [w1,w2,w3,w4]=[0.2,0.2,0.3,0.3]. The total loss measures how well the predicted airway matches the GT, with a value of 0 indicating a perfect match and a value of 1 indicating no overlap between the two. The smooth factor is added to avoid numerical instability caused by division by zero.

*E. Uncertainty Estimation methods*

**Dropout neural networks,** utilizes dropout regularization technique, it can be viewed as introducing a mask into the neural network, which randomly sets some neurons' outputs to zero. The approach involves incorporating a dropout layer at the

| **Algorithm 1** Pseudocode of uncertainty estimation of inference |
|---|
| **Require:** The trained DL model, dropout for n_drop times |
| **Ensure:** The input image X |
| 1: **for** n in range (n_drop) **do** inference : pred=model(X) |
| 2: add pred for n_drop times: predictions=append(pred) |
| 3: **end for** |
| 4: Out=predictions/n_drop |
| 5: mean(predctions) var(predctions) // Compute the mean and variance of the predicted results. |
| 6: **returns** mean, var, Out |

conclusion of every convolutional layer, ensuring its utilization during both training and testing phases. Suppose the input to a certain layer is denoted as x, when using Dropout, we introduce a corresponding mask vector m, where each element of m is either 0 or 1, indicating whether the corresponding neuron is retained or not. During training, we multiply the output of each

neuron by a retention probability p, the output of dropout H can be represented as (4):

$$H = p * m * x \quad (4)$$

$$g' = p * m * g \quad (5)$$

When considering the gradient computation during the backpropagation process in Dropout, adjustments need to be made to maintain the consistency of gradients. Assuming the gradient at a certain layer is denoted as g, and the corresponding gradient after Dropout is denoted as g'. Applying the chain rule, we can derive equation (5). Indeed, Dropout can effectively encourage the network to learn more robust and generalized feature representations. By randomly dropping out neurons during training, it reduces the excessive reliance on specific neurons, thus preventing overfitting and improving the model's generalization ability.

**Monte Carlo** [12] method is used in the inference process to estimate the model's performance with confidence. It involves performing multiple inferences on the same input, each time applying Dropout and then averaging the results. This approach provides a more accurate estimate of the model's performance by accounting for the uncertainty introduced by Dropout.

*F. Post-processing*

Maximum connectivity calculation: Connected Component Analysis (CCA) is used as a common post-processing technique in a number of state-of-the art algorithms [13]. This technique is used to isolate individual components of the segmentation output using connected neighbourhoods and label propagation. In order to filter out unwanted and disconnected segments under a threshold, the skimage.measure.label() function can label connected regions of a binary image. The return value of the function is the labelled image and the number of assigned labels, and does not change the binary image. A NumPy array is created to store the volume of each labeled region in the input image. A loop iterates through each labeled region, and uses NumPy's boolean indexing and sum() function to calculate the volume of the region. Then the volume array is sorted in ascending order and the largest region is selected and kept by indexing.

### III. EXPERIMENTS AND RESULTS

*A. Evaluation Metrics*

Predictions were made on our test set and the results were compared with the ground truth annotated by the radiologist to calculate four metrics to measure the performance of the model, namely Dice similarity coefficient (DSC), Precision, Tree Dsetected ratio (TD) and Branch Detected ratio (BD).

In this paper, we compared our baseline UNet model, our Branch-level UNet (B-UNet) and our Branch-level CE-UNet (B-CE-UNet) with five state-of-the-art models to see the performance of segmenting the pulmonary airways. As illustrated in TABLE I, in terms of performance, our B-UNet and B-CE-UNet demonstrate comparable results. Specifically, the B-UNet model achieves higher performance in metrics such as Dice Similarity Coefficient (DSC), Precision, and Tree Detected ratio (TD), attaining respective scores of 89.7%, 89.3%, and 90.1%. On the other hand, when comparing our B-UNet model with the B-CE-UNet model, the former shows a lower score in the Branch Detected ratio (BD) metric with 87.8%, while the latter achieves a score of 88.4%. However, it should be noted that the B-CE-UNet model also exhibits slightly lower performance in the DSC, Precision, and TD metrics compared to the B-UNet model. Specifically, the B-CE-UNet model achieves scores of 88.9%, 88.6%, and 89.9% in these metrics, respectively. These scores are only marginally lower than those achieved by the B-UNet model.

TABLE I. EVALUATION METRICS OF STATE-OF-THE-ART MODELS AND OURS

| Evaluation Metrics | Models | | | | | | | |
|---|---|---|---|---|---|---|---|---|
| | *Attention UNet* [14] | *WingsNet* [15] | *V-Net* [16] | *VoxResNet* [17] | *AG U-Net* [18] | *Our (Baseline)* | *Our B-UNet* | *Our B-CE-UNet* |
| DSC | 0.840±0.087 | 0.860±0.120 | 0.859±0.034 | 0.858±0.063 | 0.827±0.222 | 0.882±0.042 | 0.897±0.034 | 0.889±0.031 |
| Precision | 0.849±0.098 | 0.867±0.043 | 0.818±0.070 | 0.783±0.098 | 0.725±0.289 | 0.890±0.044 | 0.893±0.041 | 0.886±0.048 |
| TD | 0.851±0.019 | 0.790±0.111 | 0.35±0.098 | 0.331±0.102 | 0.635±0.308 | 0.807±0.111 | 0.901±0.070 | 0.899±0.076 |
| BD | 0.756±0.023 | 0.805±0.125 | 0.342±0.091 | 0.298±0.099 | 0.701±0.333 | 0.750±0.138 | 0.878±0.096 | 0.884±0.100 |

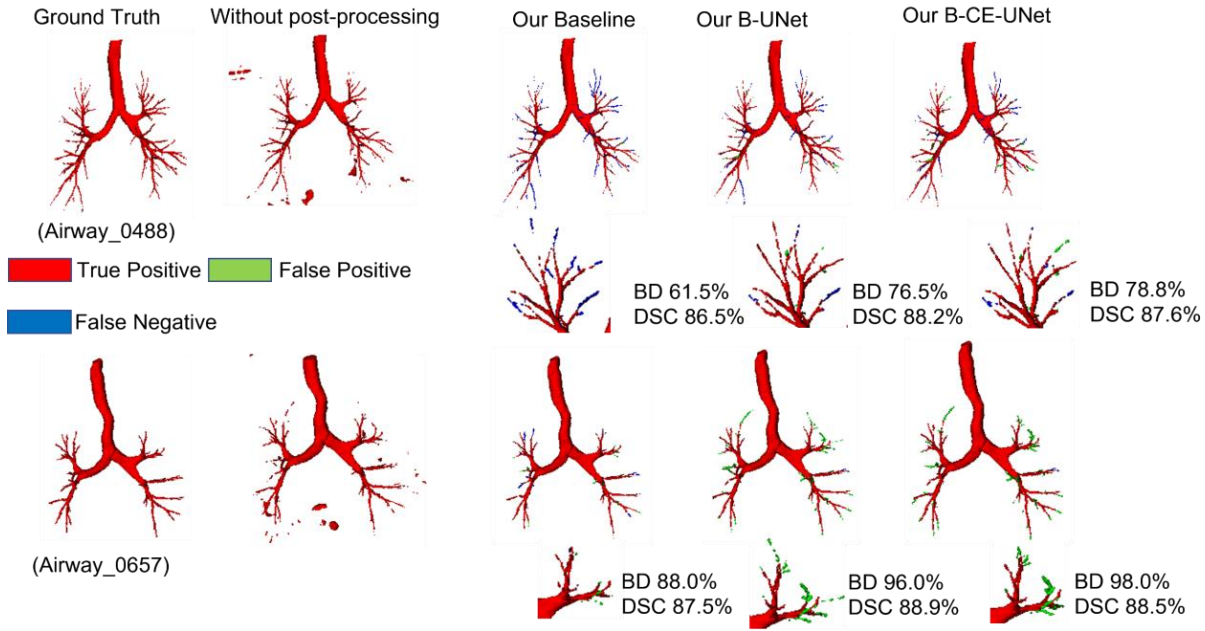

Fig. 5 Visualisation of segmentation results. *Case Airway_0488 is a sample with low BD score but high Dice, Case Airway_0657 is a sample with high BD but low Dice. The blue and green markings represent false negative (FN) and false positive (FP), respectively.

However, when comparing B-UNet and B-CE-UNet with baseline 3D UNet model, it becomes evident that the choice of loss functions has a significant impact. The baseline model solely incorporates the fundamental Dice Loss (dice similarity coefficient) and BCE Loss (binary cross entropy), which results in relatively favorable performance in terms of DSC and Precision metrics. However, the absence of specialized loss functions specifically designed for the segmentation of lung airways, such as Branch Loss and Central Line Loss (as described in Sections C and D of the METHODS chapter), leads to poorer performance of the baseline model in terms of the TD and BD metrics, achieving only 80.7% and 75.0%, respectively.

Furthermore, we observed that both our proposed models, B-UNet and B-CE-UNet, outperformed state-of-the-art DL models in terms of the four evaluation metrics: DSC, Precision, TD, and BD. Additionally, our baseline model also exhibited favorable performance in terms of DSC and Precision. These findings provide evidence that the 3D UNet serves as a suitable baseline model and can be preferred for developing image segmentation tasks involving lung airways and similar structures, such as branching, vascular textures, and other complex shapes. In TABLE I, the red font represents the best-performing metrics, while the blue font represents the second-best metrics.

### B. Quantitative Results

With the baseline model of 3D UNet, we compared our baseline UNet, B-UNet and B-CE-UNet visually. As showed in Fig.5., in the case of **Airway_0488**, the blue regions in the airway represent accurately predicted segments (false negatives, FN). Both B-UNet and B-CE-UNet models exhibited better overall performance, resulting in fewer blue regions compared to the baseline model. Upon closer examination of the magnified detail images, it is evident that B-UNet and B-CE-UNet models demonstrate more accurate predictions at the ends of the airways. However, in the case of Airway_0488, the predicted BD value is relatively low, indicating poorer capability in identifying unmarked regions at the end of the airway. This observation is further supported by the presence of green regions in the detail images, representing areas where the airway was not accurately segmented.

In contrast to the Airway_0488 case, we identified the opposite scenario in the case of **Airway_0657**. In Airway_0657, all three models consistently performed well in terms of the BD metric but exhibited lower performance in terms of DSC. From Fig. 5, it is apparent that the models with fewer blue regions demonstrate higher accuracy in their predictions, closely resembling the Ground Truth (GT). Additionally, the higher BD scores in this case indicate strong predictive capabilities of the models for the terminal branches of the airway. However, the presence of a few green regions suggests instances of incorrect predictions (false positives, FP) made by the models. In this context, the higher occurrence of green regions signifies the models' ability to predict airway terminal branches that were not annotated by human experts, highlighting their capability to identify subtle branching patterns that are difficult for manual annotation. From these two examples, we have discovered that having a higher DSC value than BD does not necessarily imply that the model achieves the desired performance. A high DSC value only indicates how closely the model's predictions align with the Ground Truth (GT). In the context of lung airway segmentation, when BD is higher than DSC, it does not necessarily imply poor model performance. A higher BD score suggests that the model's predictions may not align perfectly with the Ground Truth, but it can effectively predict unannotated regions or areas that are imperceptible or overlooked by human observers. Additionally, models with higher BD scores often demonstrate stronger predictive capabilities for the terminal branches of the airway. In summary, a higher DSC value indicates closer agreement with the GT, while a higher BD value indicates better performance in predicting the terminal branches of the airway and unannotated regions. Furthermore, in Fig. 5, we also visualized the prediction results of the two examples without undergoing post-processing. It is evident that the predictions without post-processing exhibit noticeable False Positives, which are nodules disconnected from the airway or abnormal nodules connected to the airway. These nodules differ from the false positives at the terminal branches of the airway predicted by the model (which are, in fact, unannotated airways). Therefore, it is necessary to apply a filtering process based on the calculation of the maximum connectivity rate of the airway to remove these abnormal values that do not belong to the airway, thus improving the accuracy of the model's predictions.

## IV. CONCLUSION

In this paper, we proposed a baseline UNet, Branch-level UNet (B-UNet) and Branch-level CE-UNet for 3D pulmonary airway segmentation. Branch Loss and Central Line Loss are specifically introduced in this paper, with a novel network structure which included dense atrous convolution (DAC) and residual multi-kernel pooling (RMP). The performance of our B-UNet and B-CE-UNet is higher than the baseline model and other 5 state-of-the-art models. Our model performs well in this specific task, as it can not only predict results highly similar to the GT but also identify unannotated airways and the terminal airways. This capability holds significant importance in assisting diagnosis and prognosis-related lung diseases.Keep your text and graphic files separate until after the text has been formatted and styled. Do not use hard tabs, and limit use of hard returns to only one return at the end of a paragraph. Do not add any kind of pagination anywhere in the paper. Do not number text heads-the template will do that for you.